%% file: main.tex
\documentclass[letterpaper,10pt,conference]{ieeeconf}
\IEEEoverridecommandlockouts
\usepackage{cite}
\usepackage{amsmath,amssymb,amsfonts}
\usepackage{graphicx}
\usepackage{booktabs}
\usepackage{array}
\usepackage{tabularx}
\usepackage{url}
\usepackage{xcolor}
\usepackage{flushend}
\graphicspath{{figures/}}

\newcommand{\interface}{\mathcal{I}}
\newcommand{\sat}{\operatorname{sat}}
\newcommand{\PFull}{\mathrm{P\text{-}Cal}}
\newcommand{\PDirect}{\mathrm{P\text{-}Limit}}
\newcommand{\PNoFilter}{\mathrm{P\text{-}NoFilt}}
\newcommand{\VBase}{\mathrm{V\text{-}Base}}
\newcommand{\VDam}{\mathrm{V\text{-}Match}}

\input{three_task_results.tex}

\begin{document}

\title{\LARGE \bf Same Label, Different Interface: \\ Beyond Action-Space Labels in Robot Learning}

\author{Arshia Sangwan$^{1}$, Zhongyang Zhang$^{2}$
\thanks{$^{1}$New York University, $^{2}$University of California San Diego}
\thanks{Corresponding author: Arshia Sangwan, \texttt{as22034@nyu.edu}.}}

\maketitle

\begin{abstract}
Robot-learning studies often compare position and velocity actions as though each label identified a single experimental treatment, yet the label alone does not determine what the robot executes. Before affecting robot motion, bounded policy output passes through filtering, physical scaling, reference construction, feedback, timing, and effort limits. We study how these omitted properties affect attribution under one shared learner and training budget. Across three contact-rich Factory tasks (PegInsert, GearMesh, and NutThread), we compare five joint-space interfaces, with ten independently trained policies per condition. We first remain within the same position-feedback law and vary the executed interface: substituting only the physical action scale as the directly modified component changes sample-median evaluation success by 0.645--1.000 across tasks, while removing an exponential moving average produces task-dependent changes. We then move to a cross-law comparison and choose a velocity scale that exactly matches the position interface's local raw-torque command gain. This calibration raises the sample median on GearMesh and NutThread, with no median change on PegInsert. Yet clear residual sample-median gaps remain on PegInsert and GearMesh, while the NutThread interval includes zero. Thus, even this explicit equality leaves feedback and held-action dynamics unmatched and does not establish position--velocity trained-outcome equivalence. Action-interface comparisons should therefore report the executed pathway and state the physical quantity they match.
\end{abstract}

\section{Introduction}

Two systems can both expose ``joint-position actions'' while mapping the same normalized policy output to very different target changes. One may request a small calibrated joint increment. Another may request an increment based on the distance from the nominal pose to each joint limit. The label is accurate in both cases, yet it omits a physical difference that can change the robot trajectories used for learning.

At each policy step, the system samples $\tilde a_t\sim\pi_\theta(\cdot\mid o_t)$ from observation $o_t$ and bounds it as $a_t=\mathcal B(\tilde a_t)=\operatorname{clip}(\tilde a_t,-1,1)$. The implemented pathway then filters the bounded action, applies command coordinates and physical scale, constructs and holds a reference, evaluates feedback, and limits motor torque. These operations determine how a normalized output changes the robot state. Position, velocity, torque, and impedance name only selected parts of that pathway.

\begin{figure*}[t]
\centering
\includegraphics[width=\textwidth]{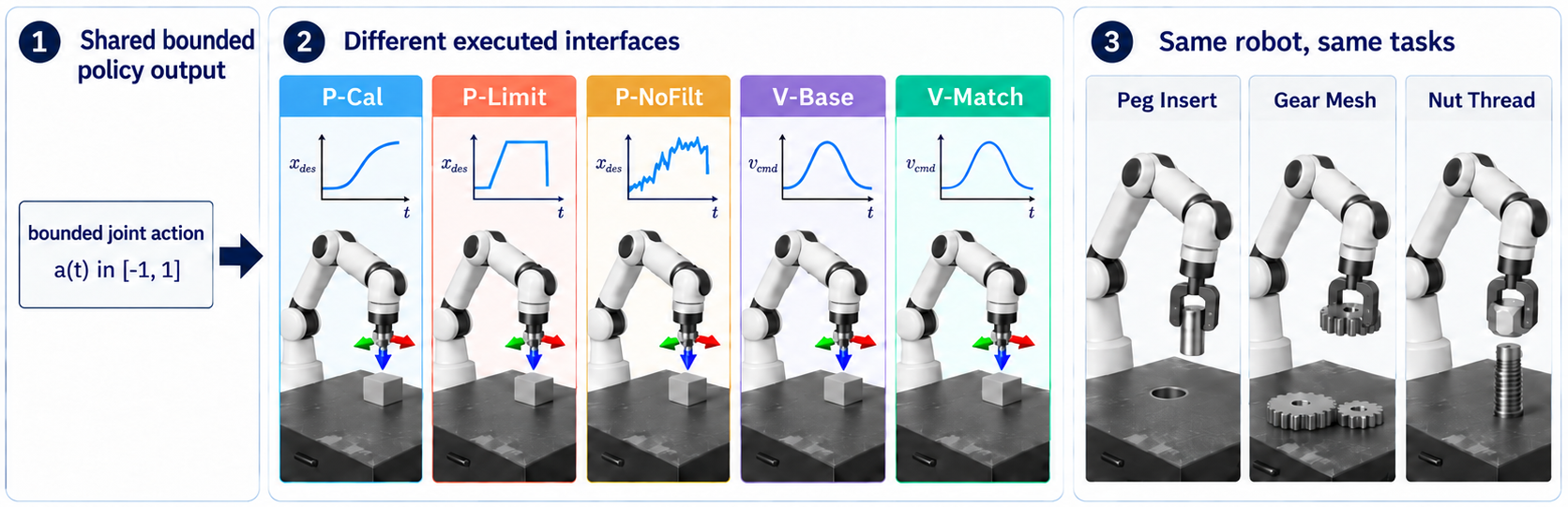}
\caption{Shared bounded policy output does not determine the executed interface. P-Cal, P-Limit, and P-NoFilt form the within-position-law counterexample. V-Base provides the unmatched velocity baseline, while P-Cal and V-Match form the cross-law command-gain test. The small command curves are schematic, and all conditions use the same robot and tasks.}
\label{fig:overview}
\end{figure*}

\begin{figure*}[t]
\centering
\includegraphics[width=\textwidth]{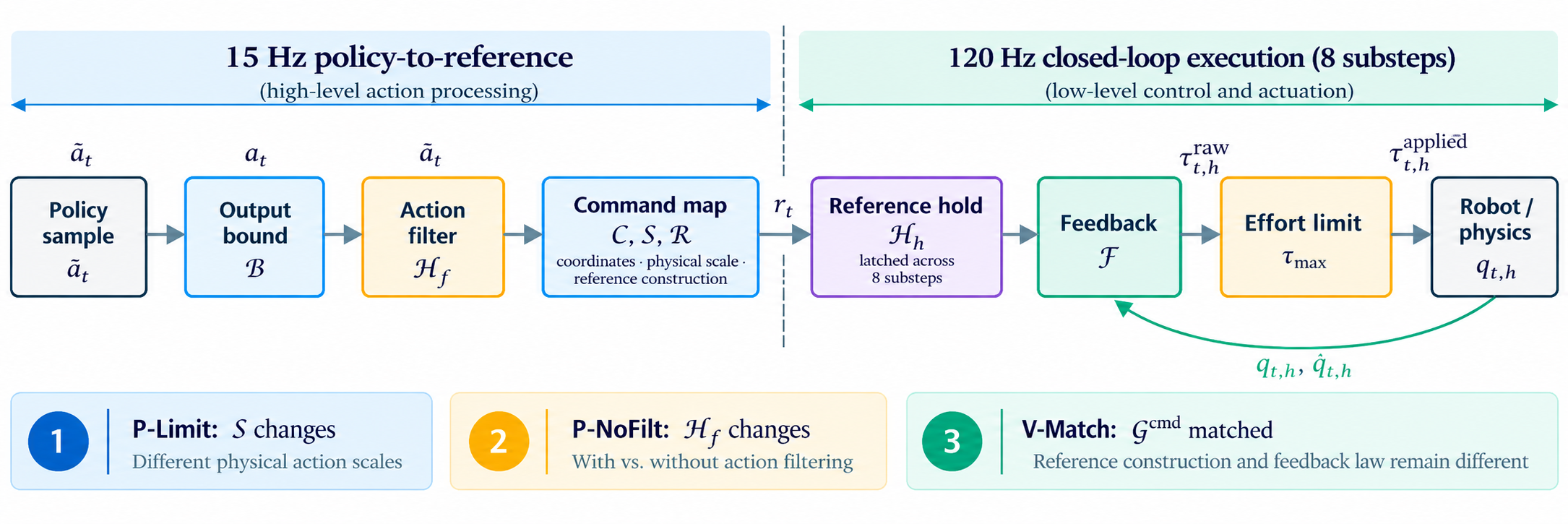}
\caption{The executed interface separates the 15~Hz policy-to-reference pathway from eight 120~Hz closed-loop physics substeps. The lower boxes locate the directly changed or explicitly matched component for P-Limit, P-NoFilt, and V-Match.}
\label{fig:interface}
\end{figure*}

Figure~\ref{fig:overview} introduces the tested interfaces and tasks. The three position conditions show how one nominal label can contain different physical scales or temporal processing. The velocity conditions test a stronger cross-law matching hypothesis. Figure~\ref{fig:interface} shows the actual order of operations. Filtering precedes coordinate mapping and reference construction. The resulting reference is latched once per policy decision, while feedback and effort limiting are evaluated on each physics substep.

Existing work shows that action representation, temporal persistence, gains, and controller choice affect learning \cite{varin2019comparison,martinmartin2019vices,metelli2020frequency,aljalbout2024actionspace,bronars2026tune,esser2025actionspace,azimi2026benchmarking}. The unresolved experimental-design question is what an observed difference can be attributed to when a nominal label leaves several of these properties unspecified.

We study this question through controlled substitutions on Factory PegInsert, GearMesh, and NutThread. We first preserve the position feedback law while changing physical action scale or exponential moving-average processing. We then change to velocity feedback and match a precisely defined local raw-torque command gain. This calibration equalizes the action-driven raw-torque coefficient while leaving the feedback terms and held-action state evolution different. The required matching level follows from the attribution being made. A position-versus-velocity claim needs different controls than a claim about physical scale. Our estimand is interface substitution under one shared learner configuration and budget. Separate retuning answers the complementary optimization question. The paper makes three contributions.

\begin{enumerate}
\item \textbf{An attribution framework for action interfaces.} We separate the operations between bounded policy output and applied torque, which makes each controlled substitution and its licensed interpretation explicit.
\item \textbf{Controlled within-law evidence.} With position feedback fixed, directly changing only the physical action scale produces large outcome differences across all three tasks. Removing the moving average produces task-dependent changes.
\item \textbf{A controlled test of local command-gain matching.} We match the local raw-torque command gain across position and velocity feedback. The calibration changes some observed outcomes but neither matches the remaining feedback dynamics nor equalizes the observed trained-policy outcomes.
\end{enumerate}

The methodological rule is direct: state what is physically matched and report what remains different. We use one learner and one training budget so that interface substitution remains the experimental treatment. Retuning the learner for each interface asks the complementary question of what performance each interface can reach after separate optimization. Our evidence covers one robot, one simulator family, and three contact tasks within the same stack.

\section{Related Work}

Prior work establishes that action and controller design affect robot learning. Studies compare position, velocity, torque, pose, and joint commands across manipulation settings \cite{varin2019comparison,aljalbout2024actionspace,esser2025actionspace,azimi2026benchmarking}. Other work isolates absolute versus incremental commands, policy initialization, behavior between policy decisions, and temporal persistence \cite{esser2025actionspace,pmlr-v306-feng26ab,metelli2020frequency}. Controller-gain studies show that learnability and the benefit of separate learner tuning depend on the gain regime \cite{bronars2026tune}. Variable-impedance and rotation-action studies further demonstrate that command semantics matter beyond a coordinate name \cite{martinmartin2019vices,bogdanovic2020variable,yang2022variable,schuck2026so3}.

Prior studies establish the empirical importance of alternative action representations, controllers, gains, and temporal choices. We focus on the attribution enabled by such comparisons. We expose a consequential scale difference hidden beneath one position-action label, then construct a local raw-torque command-gain equality across two feedback laws and test the limit of that equality. E\ss er et al. and Feng et al. study broad spatial and temporal action-space design \cite{esser2025actionspace,pmlr-v306-feng26ab}. Bronars et al. study controller gains and the benefit of separate learner tuning \cite{bronars2026tune}. Our controlled substitutions instead ask what physical treatment a nominal label identifies under one fixed learner and what conclusion the comparison supports.

Classical impedance and operational-space control distinguish command coordinates, feedback, and closed-loop dynamics \cite{hogan1985impedance,khatib1987operational}. We use standard control equations to identify the terms that remain different after the local gain is matched. Factory and Isaac Lab provide one robot and simulator family across distinct assembly tasks \cite{narang2022factory,mittal2025isaaclab}. This setting lets us repeat the same within-task intervention without changing the surrounding software stack.

\section{Definitions}

\subsection{What the interface contains}

For the deterministic actuation pathway from a policy sample to applied torque, we use the ordered description
\begin{equation}
\interface=(\mathcal B,\mathcal H_f,\mathcal C,\mathcal S,\mathcal R,
\mathcal H_h,\mathcal F,\Delta t,\boldsymbol\tau_{\max}),
\label{eq:interface}
\end{equation}
Here $\mathcal B$ bounds the policy output, $\mathcal H_f$ filters the action history, $\mathcal C$ specifies command coordinates, $\mathcal S$ sets physical scale, $\mathcal R$ constructs the reference, $\mathcal H_h$ holds that reference, $\mathcal F$ is the feedback law, $\Delta t$ is the policy period, and $\boldsymbol\tau_{\max}$ gives joint-effort limits. The notation follows implementation order rather than grouping all temporal operations into one block. Learner architecture, optimization, rewards, and randomization lie outside this actuation tuple. Those settings remain fixed. Removing the filter also changes the semantic content of the previous-action feature, which we report as a coupled consequence of the implemented temporal pathway. The executed chain is
\begin{equation}
\begin{aligned}
\tilde a_t&\xrightarrow{\mathcal B}a_t\xrightarrow{\mathcal H_f}\bar a_t
\xrightarrow{\mathcal C,\mathcal S}c_t^{phys}\xrightarrow{\mathcal R}r_t,\\
r_t&\xrightarrow{\mathcal H_h,\mathcal F}\tau^{raw}_{t,0:L-1}
\xrightarrow{\boldsymbol\tau_{\max}}\tau^{applied}_{t,0:L-1}.
\end{aligned}
\label{eq:chain}
\end{equation}
Here $c_t^{phys}$ is the scaled command in the interface's physical command coordinates, $r_t$ is the latched reference, and $L$ is the number of physics substeps per decision. Superscripts ``raw'' and ``applied'' distinguish torque before and after effort limiting. A position or velocity label usually identifies the reference or feedback category. It does not specify Eq.~\eqref{eq:interface}.

\subsection{Fixed-budget outcomes and equivalence}

Each independently trained policy is one statistical sample. Its continuous deterministic evaluation success fraction at the common training budget is the primary outcome. We also report the number of policies at or above 0.5 as an interpretable descriptive summary. The 0.5 value is not an independently validated competence threshold. The experiment does not measure learning speed, sample efficiency, or the best result after interface-specific tuning.

We distinguish three levels of equality. Command-gain equality means that the action-driven raw-torque coefficient is equal. Closed-loop equivalence would require equal state and applied-torque responses over the held action. Trained-outcome equivalence would preserve the policy-outcome distribution under the shared learner and budget. Our analytic intervention establishes the first level. The finite policy sample tests whether that condition equalizes the observed outcomes, not whether two population distributions are identical.

\section{Interfaces and Controlled Changes}

All conditions command the seven Franka joints, use $a_t=\operatorname{clip}(\tilde a_t,-1,1)$, and share the final effort limit. Four conditions use the exponential moving average
\begin{equation}
\bar a_t=(1-\alpha)\bar a_{t-1}+\alpha a_t,
\qquad \alpha=0.2.
\label{eq:ema}
\end{equation}
For the unfiltered position condition, $\bar a_t=a_t$. The incremental position interface constructs a target once per policy decision,
\begin{align}
q_t^\star &= q_{t,0}+S_P\bar a_t,\label{eq:ptarget}\\
\tau^{P,raw}_{t,h} &=K_p(q_t^\star-q_{t,h})-K_d\dot q_{t,h},\\
\tau^{P,applied}_{t,h}&=\sat_{\boldsymbol\tau_{\max}}(\tau^{P,raw}_{t,h}),
\label{eq:plaw}
\end{align}
where $q_{t,0}$ is the live joint configuration at the start of decision $t$. The interface holds $q_t^\star$ for $L=8$ simulator substeps. We use $K_p=80$ and $K_d=4$ on each controlled joint. The velocity interface constructs
\begin{align}
\dot q_t^\star &= \frac{S_V}{\Delta t}\bar a_t,\label{eq:vtarget}\\
\tau^{V,raw}_{t,h} &=K_v(\dot q_t^\star-\dot q_{t,h}),\\
\tau^{V,applied}_{t,h}&=\sat_{\boldsymbol\tau_{\max}}(\tau^{V,raw}_{t,h}),
\label{eq:vlaw}
\end{align}
where $K_v=[40,40,40,40,4.5977,4.5977,4.5977]$. Products are joint-wise. These gains were fixed independently of learning outcomes. $K_p=80$ and $K_d=4$ are the measured default stiffness and damping in the committed Franka asset record. The velocity gains are the effort limits divided by rated joint speeds, giving $87/2.175=40$ for joints 1--4 and $12/2.610=4.5977$ for joints 5--7.

Table~\ref{tab:interfaces} summarizes the five conditions. The physical action scale is calibrated independently for each task except for the shared P-Limit vector. $\PFull$ is the calibrated reference. P-Limit is intentionally a constructive witness to label under-specification, not a proposed performance baseline. $\PDirect$ retains its law, gains, incremental target, moving average, rate, and effort limit but replaces the calibrated scale with a per-joint limit-distance scale. Each component is the larger distance from the nominal configuration to either joint limit. Because Eq.~\eqref{eq:ptarget} remains a live increment, this vector sets requested displacement per decision rather than an absolute lower-to-upper joint range. $\PNoFilter$ removes the moving average. This changes the temporal transformation and the meaning of the previous-action observation while preserving its dimensionality and the rest of the learner. $\VBase$ uses $S_V=S_P$ before conversion to desired velocity by $1/\Delta t$. $\VDam$ uses a gain-derived velocity scale that matches the local raw-torque command gain defined below.

\begin{table}[t]
\centering
\caption{Five tested joint-space interfaces.}
\label{tab:interfaces}
\footnotesize
\renewcommand{\arraystretch}{1.08}
\setlength{\tabcolsep}{1.1pt}
\begin{tabular}{lccccc}
\toprule
Interface & Feedback & Target & \shortstack{Physical\\scale} & Temporal & Role\\
\midrule
$\PFull$ & position & increment & calibrated & EMA & calibrated reference\\
$\PDirect$ & position & increment & limit dist. & EMA & scale substitution\\
$\PNoFilter$ & position & increment & calibrated & none & EMA removal\\
$\VBase$ & velocity & velocity & $S_V=S_P$ & EMA & velocity reference\\
$\VDam$ & velocity & velocity & gain-derived & EMA & command-gain test\\
\bottomrule
\end{tabular}
\end{table}

The five interfaces support four controlled comparisons rather than a ranking of unrelated controllers. Table~\ref{tab:matched} separates the directly modified component from downstream consequences. In the scale substitution, raw torque, saturation, trajectories, and contacts may change because $S_P$ changes. They are not independently manipulated factors.

\begin{table*}[t]
\centering
\caption{Four controlled interface comparisons.}
\label{tab:matched}
\footnotesize
\renewcommand{\arraystretch}{1.12}
\setlength{\tabcolsep}{3.2pt}
\begin{tabularx}{\textwidth}{
    p{0.15\textwidth}
    p{0.12\textwidth}
    p{0.30\textwidth}
    p{0.11\textwidth}
    X
}
\toprule
Comparison & Direct change & Held fixed & Explicit match & Question\\
\midrule
$\PFull$ vs. $\PDirect$
& physical scale $S_P$
& P target, feedback, gains, EMA, rates, limits; learner architecture and optimization
& none
& Can one position label hide a consequential scale change?\\

$\PFull$ vs. $\PNoFilter$
& EMA pathway
& P target, scale, feedback, rates, limits; learner architecture and optimization
& none
& What changes when the implemented EMA path is removed?\\

$\VBase$ vs. $\VDam$
& velocity scale $S_V$
& V target, feedback, gains, EMA, rates, limits; learner architecture and optimization
& $G_V^{cmd}=G_P^{cmd}$
& Does removing one command-gain mismatch change outcomes?\\

$\PFull$ vs. $\VDam$
& $\mathcal{R}, \mathcal{S}, \mathcal{F}$
& EMA, rates, limits; learner architecture and optimization
& $G_V^{cmd}=G_P^{cmd}$
& Is command-gain equality sufficient?\\
\bottomrule
\end{tabularx}
\end{table*}
Removing the EMA changes the temporal operator as a whole. Relative to $\PFull$, it removes temporal averaging and increases the local raw-torque derivative with respect to the current bounded action from $\alpha K_pS_P$ to $K_pS_P$. With $\alpha=0.2$, this current-action sensitivity increases by a factor of five. The previous-action feature also changes from the previous filtered action to the previous bounded action. The comparison therefore isolates EMA processing as implemented, not memory or instantaneous sensitivity alone.

\subsection{Choosing the task-specific position scale}

We choose the position scale without training a policy. For each task, eight resets of 128 environment instances give 1024 states. Generator seed 700 produces 64 componentwise-uniform directions in $[-1,1]^6$ for Cartesian commands and $[-1,1]^7$ for joint commands. The directions are not normalized. At every state--direction pair, we measure the RMS one-step task-space displacement
\begin{equation}
\begin{aligned}
d_{task}(\delta q;J)&=
\sqrt{\lVert J_v\delta q\rVert_2^2+
\lambda_x^2\lVert J_\omega\delta q\rVert_2^2},\\
\lambda_x&=0.02/0.097~\mathrm{m/rad}.
\end{aligned}
\label{eq:pam}
\end{equation}
Here $\delta q$ is a joint displacement, while $J_v$ and $J_\omega$ are the translational and rotational geometric-Jacobian blocks at a reset state. Factory scales the native Cartesian action by 0.02 m in translation and 0.097 rad in rotation. Their ratio gives $\lambda_x=0.02/0.097$ m/rad, which converts angular motion to the translation-weighted metric. Let $J=[J_v^\top,J_\omega^\top]^\top$ and let $\delta x$ denote the scaled pose command. We use Factory's damped least-squares form $\delta q=J^\top(JJ^\top+\lambda_{\mathrm{DLS}}^2I_6)^{-1}\delta x$, where $I_6$ is the task-space identity and $\lambda_{\mathrm{DLS}}=0.1$ is the factory DLS value used in our Isaac Lab 2.3.0 setup. Because the joint-side metric is linear in a uniform scale, the closed form is $S_P=d_{\mathrm{RMS}}^{cart}/d_{\mathrm{RMS}}^{joint}(S=1)$. This gives each task a position scale based on its reset states and the stated weighting. No policy, reward, or learning result is used. Table~\ref{tab:scales} reports radians per unit bounded action per decision. The desired velocity is $S_V\bar a_t/\Delta t$, and the P-Limit vector is shared across tasks. These values are action scales, not geometric task lengths.

\begin{table}[t]
\centering
\caption{Physical scales in radians per decision.}
\label{tab:scales}
\footnotesize
\renewcommand{\arraystretch}{1.08}
\setlength{\tabcolsep}{1.8pt}
\begin{tabular}{lccc}
\toprule
Scale (rad/decision) & PegInsert & GearMesh & NutThread\\
\midrule
$S_P$ for $\PFull,\PNoFilter$ & 0.034777 & 0.034186 & 0.034035\\
$S_V$ for $\VBase$ & 0.034777 & 0.034186 & 0.034035\\
$S_V^{match}$, joints 1--4 & 0.004637 & 0.004558 & 0.004538\\
$S_V^{match}$, joints 5--7 & 0.040341 & 0.039656 & 0.039481\\
$S_P$ for $\PDirect$, joints 1--4 & \multicolumn{3}{c}{$[2.8973,2.3318,2.8973,2.7402]$}\\
$S_P$ for $\PDirect$, joints 5--7 & \multicolumn{3}{c}{$[2.8973,3.0545,3.6383]$}\\
\bottomrule
\end{tabular}
\end{table}

\subsection{Matching the local raw-torque command gain}

For a fixed robot state and moving-average history, we define the local raw-torque command gain as the sensitivity before final effort limiting:
\begin{align}
G_P^{cmd}&=\frac{\partial\tau^{P,raw}}{\partial\bar a_t}=K_pS_P,\\
G_V^{cmd}&=\frac{\partial\tau^{V,raw}}{\partial\bar a_t}=K_vS_V/\Delta t.
\label{eq:gains}
\end{align}
V-Match chooses the velocity scale analytically,
\begin{equation}
S_V^{\mathrm{match}}=K_v^{-1}K_pS_P\Delta t,
\qquad G_V^{cmd}=G_P^{cmd},
\label{eq:dam}
\end{equation}
so the local raw-torque command gains are equal joint by joint. This equality concerns the action-driven contribution to raw torque. It does not match the state-feedback terms, state evolution during the held action, reference resets, or operation at the effort limit. The raw-torque derivative does not depend on whether the downstream effort limit is active. For the current bounded action, $\partial\tau^{P,raw}/\partial a_t=\alpha G_P^{cmd}$ and $\partial\tau^{V,raw}/\partial a_t=\alpha G_V^{cmd}$. The moving-average position and velocity interfaces share $\alpha=0.2$, so these raw coefficients are also equal. The applied-torque derivatives equal them only where effort limiting is inactive. Relative to the Gaussian sample, the derivative additionally contains the output-clip Jacobian $D_{\mathrm{clip}}(\tilde a_t)$, which is one for an unclipped coordinate and zero beyond the clipping interval. The scale is chosen without a learning result.

Equation~\eqref{eq:dam} establishes the equality analytically. A numerical PegInsert diagnostic evaluates 528 conditions per interface: 12 visited state batches, seven axis and four seeded random directions, and four magnitudes. Each condition averages a centered positive/negative difference over 32 environments. For either velocity interface, 484 conditions contain at least one effort-limited joint, leaving 44 fully observable conditions. The latter comprise 28 axis conditions and 16 random-direction conditions. The 28 axis conditions form four complete seven-axis rows, which explains the separate count of four usable rows. On those rows, measured and analytic coefficients agree within $3.09\!\times\!10^{-5}$ relative error. The diagnostic validates the raw-torque coefficient where it is observable. It does not match applied torque or closed-loop behavior.

\section{Experimental Method}

\subsection{Tasks and controls}

We use Factory PegInsert, GearMesh, and NutThread with one Franka Panda in Isaac Sim 5.1.0.0 and Isaac Lab 2.3.0. Peg insertion emphasizes translational alignment, gear meshing changes contact geometry, and nut threading adds rotational progression. Physics runs at 120 Hz and the policy at 15 Hz. The PhysX configuration uses 192 position iterations, one velocity iteration, 5 mm contact offset, zero rest offset, and static and dynamic friction of 1.0. The common torque limits are $[87,87,87,87,12,12,12]$ N\,m. Within each task, the five interfaces share the robot, simulator, observation layout, reward, episode structure, randomization, assets, solver, rates, torque limits, GPU family, learner configuration, and training budget. The previous-action value differs only as the coupled consequence of removing EMA described above. Every comparison remains within the task. PegInsert is a contact-rich simulated benchmark, not evidence of nominal geometric insertion precision.

As a descriptive task check, three NutThread policies trained with the unmodified Factory task-space interface achieve deterministic successes of 0.8594, 0.8125, and 0.9141. That interface includes Factory's native one-way rotation prior. These policies show that the task and training setup can produce successful behavior. They are neither joint-interface comparators nor an upper bound and are excluded from every primary cell.

\subsection{Learner, seeds, and deterministic evaluation}

All five interfaces use the same Isaac Lab \texttt{rl\_games} implementation and profile \cite{mittal2025isaaclab}, implementing proximal policy optimization (PPO) \cite{schulman2017ppo}. The policy network has a two-layer, 1024-unit long short-term memory (LSTM) module before a 512--128--64 exponential linear unit (ELU) multilayer perceptron, with concatenated recurrent input and layer normalization. The profile also uses an asymmetric central-value critic with the same recurrent and MLP widths on the privileged state. Inputs, values, and advantages are normalized. The continuous Gaussian has a learned, nonfixed log standard deviation initialized at zero ($\sigma=1$). Gaussian samples $\tilde a_t$ are clipped to $[-1,1]$ before filtering. Thus, action scale changes the physical consequence of the same normalized action parameterization.

Both the actor observation and asymmetric critic state include a seven-dimensional previous-action feature in every primary condition. In the four EMA conditions, this is the previous filtered normalized action. In $\PNoFilter$, it is the previous bounded action. Removing the EMA therefore changes both temporal processing and the semantic value given to the actor and critic, while dimensionality and the rest of the learner remain fixed. The recurrent policy state can also represent action history.

Training uses 128 actors, horizon 128, minibatches of 512, four mini-epochs, and 200 epochs, for 3,276,800 environment steps per policy. The adaptive learning rate starts at $10^{-4}$ with a Kullback-Leibler (KL) threshold of 0.008. We use discount 0.995, generalized advantage estimation (GAE) 0.95, PPO clip 0.2, entropy coefficient zero, and gradient norm 1. Every task-interface cell contains ten independently trained policies, using seed labels 0--9, on NVIDIA L40S GPUs. We evaluate each epoch-200 checkpoint.

Evaluation is deterministic with disjoint seed 900, 128 environments, and 256 policy steps. We use a project-specific geometric success criterion derived from Factory's position and depth thresholds. Planar base-position error must be below 2.5 mm. Signed axial displacement must fall below 4\% of hole height for PegInsert, 5\% of fixed gear-peg height for GearMesh, or 37.5\% of thread pitch for NutThread. For NutThread, this evaluation omits the yaw test used by Factory's training-time success signal. We therefore do not call it unqualified Factory success.

The 256-step evaluation window exceeds the native PegInsert horizon of 150 policy steps, while GearMesh and NutThread have horizons of 300 and 450 steps, respectively. At 15 Hz, PegInsert resets after step 149, but the evaluator retains its success-seen flag for the remaining 107 steps. Since the geometric criterion is checked after each environment step, the timeout-step check observes the reset state, which cannot satisfy the criterion. Under the configured pose and noise bounds, the reset peg origin remains at least 26.6 mm above the socket origin, whereas success requires it to be below 1 mm. PegInsert success therefore indicates that the criterion was met at least once within the full evaluation window, which can span two episodes, rather than within a single native episode. GearMesh and NutThread do not reset during evaluation. For each trained policy, we report the fraction of 128 environments satisfying the task-specific geometric criterion at least once within this window. Comparisons remain within task, with one final checkpoint and one deterministic evaluation per training seed across all 15 task-interface conditions.

\subsection{Measurements and comparisons}

The statistical unit is one independently trained policy. For each cell, we show every policy-level success fraction, the median, and the number of policies at or above the descriptive 0.5 threshold. The continuous outcome carries every primary comparison. The 128 evaluation environments estimate one trained policy's outcome and are not training replicates.

Seed labels 0 through 9 are reused across conditions, but they do not create common random streams across interfaces. We therefore treat the policies in each condition as independent samples. For each task we compute
\begin{align}
\Delta_{\mathrm{scale}} &= M(\PFull)-M(\PDirect),\\
\Delta_{\mathrm{EMA}} &= M(\PFull)-M(\PNoFilter),\\
\Delta_{\mathrm{match}} &= M(\VDam)-M(\VBase),\\
\Delta_{\mathrm{gap}} &= M(\PFull)-M(\VDam),
\end{align}
Here each reported effect is the difference between the two sample medians. A 95\% percentile bootstrap interval independently resamples the ten policies in each condition. These policy-resampling bootstrap intervals summarize uncertainty at modest $N$ rather than prove a population effect. They condition on each policy's fixed evaluation bank because individual environment outcomes were not retained. The 12 intervals are contrast-specific summaries and are not interpreted as a joint hypothesis-testing family.

\section{Results}

\begin{figure*}[t]
\centering
\includegraphics[width=\textwidth]{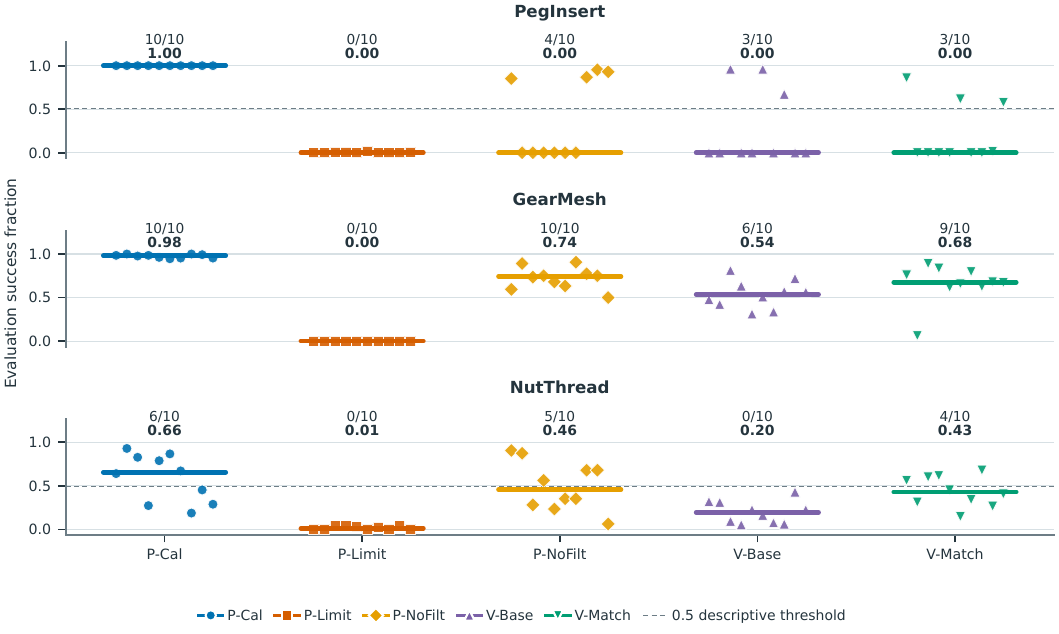}
\caption{Policy-level outcomes show strong physical-scale sensitivity and task-dependent EMA sensitivity. Each point is one independently trained policy, with $N=10$ per cell. Colored segments mark medians. Upper numbers give the descriptive threshold count, and bold lower numbers give the median. The 128 evaluation environments per policy are not training replicates.}
\label{fig:matrix}
\end{figure*}

Figure~\ref{fig:matrix} shows all 15 task-interface cells. Each dot is one independently trained policy. The colored horizontal segment is the cell median. Above each cell, the upper number is the count at or above 0.5, and the bold lower number is the median. The dashed line marks the descriptive 0.5 threshold.

\begin{figure*}[t]
\centering
\includegraphics[width=\textwidth]{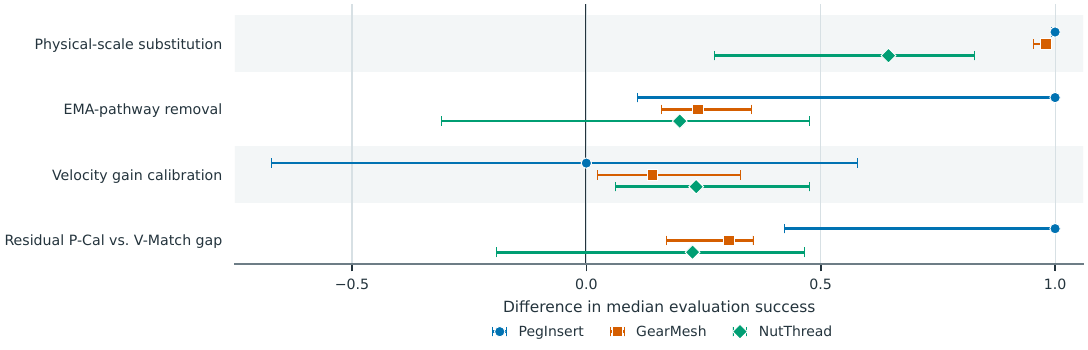}
\caption{Controlled substitutions separate physical scale, EMA processing, velocity command-gain calibration, and the remaining P-Cal versus V-Match gap. Points are differences between sample medians, and bars are 95\% independent-condition bootstrap intervals with $N=10$ per condition. Row order uses P-Cal minus P-Limit, P-Cal minus P-NoFilt, V-Match minus V-Base, and P-Cal minus V-Match.}
\label{fig:contrasts}
\end{figure*}

Figure~\ref{fig:contrasts} separates the four attribution questions. It reports differences between independent-condition sample medians with bootstrap intervals.

\subsection{Physical-scale substitution produces large within-law differences}

The first comparison tests whether two implementations sharing the same position-feedback category can induce materially different learning outcomes through physical scaling. $\PFull$ yields $\PegPFullCompetent$, $\GearPFullCompetent$, and $\NutPFullCompetent$ threshold-crossing policies on PegInsert, GearMesh, and NutThread. $\PDirect$ yields none on any task. The differences between sample medians are $\PegAuthorityDifference$, $\GearAuthorityDifference$, and $\NutAuthorityDifference$, with intervals $\PegAuthorityCI$, $\GearAuthorityCI$, and $\NutAuthorityCI$ (Fig.~\ref{fig:contrasts}). Reference construction, position feedback, gains, EMA processing, decision rate, and torque limits are fixed. The per-joint physical action scale is the only directly modified interface component. That intervention can still change downstream references, raw and applied torques, effort-limit encounters, and trajectories; those are consequences of the scale change rather than separately controlled variables.

P-Limit is a concrete alternative mapping compatible with the same joint-position label. Each joint receives a displacement scale equal to its larger nominal-to-limit distance, producing values 67 to 105 times the calibrated scale. In an auxiliary PegInsert check with $N=3$ per scale, all six policies trained at half or twice the calibrated scale cross the threshold over the tested interval from 0.0174 to 0.0696 rad per decision. This reduces the possibility that P-Cal succeeds only at one isolated scale. The check neither locates the transition to P-Limit nor separates overall magnitude from relative per-joint geometry. The primary result therefore demonstrates label under-specification rather than an optimal scale.

\subsection{Removing the EMA has task-dependent effects}

PegInsert falls from $\PegPFullCompetent$ threshold-crossing policies under $\PFull$ to $\PegPNoFilterCompetent$ under $\PNoFilter$. GearMesh retains $\GearPNoFilterCompetent$ threshold-crossing policies, while NutThread changes from $\NutPFullCompetent$ to $\NutPNoFilterCompetent$. The differences between medians are $\PegTemporalDifference$, $\GearTemporalDifference$, and $\NutTemporalDifference$, with intervals $\PegTemporalCI$, $\GearTemporalCI$, and $\NutTemporalCI$. The intervention removes temporal averaging, gives the current bounded action five times more immediate raw-torque weight, and changes the previous-action feature. It therefore identifies the effect of removing the implemented EMA pathway rather than memory alone.

\subsection{Local gain matching changes some outcomes without matching closed-loop dynamics}

The V-Base to V-Match contrast asks whether removing one identified physical mismatch changes the velocity comparison. V-Match leaves PegInsert's threshold count unchanged while raising the observed counts from $\GearVBaseCompetent$ to $\GearVDamCompetent$ on GearMesh and from $\NutVBaseCompetent$ to $\NutVDamCompetent$ on NutThread.

The continuous outcomes show the same heterogeneity (Fig.~\ref{fig:contrasts}). GearMesh has median change $\GearMatchDifference$ with interval $\GearMatchCI$, and NutThread has change $\NutMatchDifference$ with interval $\NutMatchCI$. PegInsert has change $\PegMatchDifference$ with interval $\PegMatchCI$. NutThread's V-Match median remains below 0.5 despite its positive shift. Local raw-torque command-gain calibration can materially change a comparison, but the ten-policy evidence does not support a uniform cross-task effect.
The P-Cal to V-Match contrast asks whether the local gain match is sufficient
to remove the observed position-velocity outcome difference. PegInsert retains
a median gap of $\PegResidualDifference$ with interval $\PegResidualCI$, and
GearMesh retains $\GearResidualDifference$ with interval $\GearResidualCI$.
NutThread's remaining difference is $\NutResidualDifference$ with interval
$\NutResidualCI$, which crosses zero. Thus the match removes one precise
difference without establishing closed-loop or trained-outcome equivalence.

\section{One Local Gain Leaves Feedback Unmatched}
\label{sec:mechanism}

Command-gain equality concerns only the action-driven raw-torque term, not
state feedback. Substituting the targets into the feedback laws gives:

\begin{align}
\tau^{P,raw}_{t,h}
&=\underbrace{K_pS_P\bar a_t}_{\tau^P_{cmd}}
\quad+\underbrace{\left[-K_p(q_{t,h}-q_{t,0})
-K_d\dot q_{t,h}\right]}_{\tau^P_{fb}},
\label{eq:pdecomp}\\
\tau^{V,raw}_{t,h}
&=\underbrace{K_vS_V\bar a_t/\Delta t}_{\tau^V_{cmd}}
\quad+\underbrace{\left[-K_v\dot q_{t,h}\right]}_{\tau^V_{fb}}.
\label{eq:vdecomp}
\end{align}

The position law contains the matched command term, displacement feedback from the decision-start configuration, and velocity feedback. The velocity law contains the matched command term and a different velocity-feedback term. Equation~\eqref{eq:dam} makes the command terms equal while leaving the feedback terms different.

This difference persists during the eight physics substeps between policy actions. The robot state changes after each substep, so the unequal feedback terms are repeatedly evaluated while the reference remains fixed. Equal local command gain therefore does not imply an equal held-action state transition. Reference resetting at the next decision and effort saturation introduce further differences.

\section{Discussion}

\subsection{Matching level and scientific attribution}

The appropriate matching level depends on the attribution being made. The P-Cal to P-Limit comparison holds the position-feedback law and gains, incremental target, EMA processing, rates, and torque limits fixed. Its repeated outcome difference shows that ``joint-position control'' does not identify one experimental treatment when the physical action scale changes. P-Limit is a constructive witness to that under-specification, not a recommended controller. The result does not identify an optimal scale or separate overall magnitude from relative per-joint geometry.

The P-Cal to P-NoFilt comparison changes the temporal transformation and the corresponding previous-action semantics. Its task-dependent result prevents a universal conclusion about the benefit of this EMA pathway. The V-Base to V-Match comparison removes one precisely defined local mismatch and can alter the observed learning comparison. The P-Cal to V-Match comparison shows that the resulting equality is insufficient to establish equal trained-policy outcomes.

These conclusions concern interface substitution under one shared learner configuration and training budget. Separate tuning could change the performance of any interface, but it answers a different question. The present design prevents learner changes from absorbing the physical intervention and therefore supports attribution to the substituted interface under the tested training conditions.

\subsection{Reporting the executed interface}

Reporting and matching serve different purposes. Reporting makes the treatment interpretable. Matching supports a specific attribution. A study that attributes an outcome to position versus velocity control should state the physical level at which those interfaces are compared and disclose the remaining differences. It need not make every component identical when the scientific question calls for a different design.

A reproducible action-interface description should report policy-output bounds, command coordinates, reference construction, physical scaling, temporal processing and hold behavior, feedback law and gains, policy and physics rates, actuator limits, previous-action observation semantics, and whether learner settings are shared or independently tuned. These fields form a practical checklist rather than a claim that each field is always causal. A stronger equivalence claim would also compare robot response throughout the held action, including operation at effort limits.

\section{Limitations}

The study uses one simulated Franka model, one simulator family, three Factory tasks, and one PPO configuration. It estimates fixed-budget outcomes under shared learner settings rather than performance after separate interface-specific tuning. All comparisons remain within task because rewards, success criteria, and timeout behavior differ. The NutThread task-space control includes a one-way rotation prior and serves only as a task check. Bootstrap intervals condition on a fixed evaluation bank. The local gain equality does not establish a sufficient metric for full held-action response. These conditions bound the conclusions to within-stack comparisons and do not support real-robot, sim-to-real, or universal controller claims.

\section{Conclusion}

An action-space label does not specify the executed learning interface. Physical scaling and EMA processing can change outcomes while the nominal controller category remains fixed. Matching a local raw-torque command gain removes one physical difference across feedback laws and can change the observed comparison. It still leaves their feedback response and held-action state evolution different. In robot-learning action-interface comparisons, the executed interface, not its nominal action-space label, defines the treatment being compared.

\bibliographystyle{IEEEtran}
\bibliography{refs}

\end{document}

%% file: three_task_results.tex
\newcommand{\PegPFullCompetent}{10/10}

\newcommand{\PegPNoFilterCompetent}{4/10}

\newcommand{\PegAuthorityDifference}{+1.000}
\newcommand{\PegAuthorityCI}{[+0.992,\,+1.000]}

\newcommand{\PegTemporalDifference}{+1.000}
\newcommand{\PegTemporalCI}{[+0.109,\,+1.000]}

\newcommand{\PegMatchDifference}{+0.000}
\newcommand{\PegMatchCI}{[-0.672,\,+0.578]}

\newcommand{\PegResidualDifference}{+1.000}
\newcommand{\PegResidualCI}{[+0.422,\,+1.000]}

\newcommand{\GearPFullCompetent}{10/10}

\newcommand{\GearPNoFilterCompetent}{10/10}

\newcommand{\GearVBaseCompetent}{6/10}

\newcommand{\GearVDamCompetent}{9/10}

\newcommand{\GearAuthorityDifference}{+0.980}
\newcommand{\GearAuthorityCI}{[+0.953,\,+0.992]}

\newcommand{\GearTemporalDifference}{+0.238}
\newcommand{\GearTemporalCI}{[+0.160,\,+0.352]}

\newcommand{\GearMatchDifference}{+0.141}
\newcommand{\GearMatchCI}{[+0.023,\,+0.328]}

\newcommand{\GearResidualDifference}{+0.305}
\newcommand{\GearResidualCI}{[+0.172,\,+0.355]}

\newcommand{\NutPFullCompetent}{6/10}

\newcommand{\NutPNoFilterCompetent}{5/10}

\newcommand{\NutVBaseCompetent}{0/10}

\newcommand{\NutVDamCompetent}{4/10}

\newcommand{\NutAuthorityDifference}{+0.645}
\newcommand{\NutAuthorityCI}{[+0.273,\,+0.828]}

\newcommand{\NutTemporalDifference}{+0.199}
\newcommand{\NutTemporalCI}{[-0.309,\,+0.477]}

\newcommand{\NutMatchDifference}{+0.234}
\newcommand{\NutMatchCI}{[+0.062,\,+0.477]}

\newcommand{\NutResidualDifference}{+0.227}
\newcommand{\NutResidualCI}{[-0.191,\,+0.465]}